\documentclass[3p,times,procedia]{elsarticle}
\usepackage{ecrc}
\usepackage[bookmarks=false]{hyperref}
    \hypersetup{colorlinks,
      linkcolor=blue,
      citecolor=blue,
      urlcolor=blue}

\volume{00}

\firstpage{1}

\journalname{Procedia Computer Science}

\runauth{T. Herlau et al.}

\jid{procs}

\usepackage{amssymb}

\usepackage[figuresright]{rotating}

\usepackage{amssymb}
\usepackage{wrapfig}
\usepackage{graphicx}
\usepackage{caption}
\usepackage{subcaption}
\usepackage{cleveref}

\newcommand{\m}[1]{\mathbf{#1}}
\newcommand{\mcal}[1]{\mathcal{#1}}
\newcommand{\thetat}{\tilde{\theta}}
\newcommand{\wt}{\tilde{w}}
\newcommand{\Normal}{\mathrm{Normal}}

\newcommand{\intd}[1]{\int \!{#1}\ }

\begin{document}
\begin{frontmatter}

\dochead{The 3rd International Workshop on Statistical Methods and Artificial Intelligence  \\(IWSMAI22)
	\\ March 22 - 25, 2022, Porto, Portugal}%

\title{Bayesian dropout}

\author[a]{Tue Herlau\corref{cor1}} 
\author[a]{Mikkel N. Schmidt}
\author[a]{Morten M{\o}rup}

\address[a]{Technical University of Denmark, Richard Petersens plads 21, 2800 Lyngby, Denmark}

\begin{abstract}
In the past decade, Dropout has emerged as a powerful and simple method for training neural networks preventing co-adaptation by stochastically omitting neurons.
Dropout is currently not grounded in explicit modelling assumptions which so far has precluded its adoption in Bayesian modeling.
Using Bayesian entropic reasoning we show that dropout can be interpreted as optimal inference under constraints.
We demonstrate this on an analytically tractable regression model providing a Bayesian interpretation of its mechanism for regularizing and preventing co-adaptation as well as its connection to other Bayesian techniques, and in our experiments we find that dropout can provide robustness under model misspecification.
Our framework roots dropout as a theoretically justified and practical tool for statistical modeling allowing Bayesian practitioners to tap into the benefits of dropout training.	
\end{abstract}

\cortext[cor1]{Corresponding author. Tlf.: +45-27830812.}
\end{frontmatter}

\email{tuhe@dtu.dk}

\section{Introduction}
Consider a probabilistic model of a dataset parameterized by the parameter vector $\m \theta$. Often, the dataset will contain complicated structure and require an expressive model. However, a complex model will have many settings of its parameters which are compatible with the training data, and often be \emph{misspecified}, meaning it does not correspond  to the underlying generative process. In this case, different settings of the parameters that are compatible with the training data may make different predictions on test data, and there is no guarantee the model will concentrate on parameter values with the better generalization error~\cite{Grunwald2007}.

In the past decade, Dropout has emerged as a powerful way to address similar problems in neural network training~\cite{Hinton2012}. Dropout stochastically perturbs (typically by setting to zero) parts of the internal representation during training. This prohibit weights to co-adapt to each other, and has been shown to outperform other state-of-the-art regularization methods~\cite{Hinton2012}.

The success of dropout has been attributed to a variety of mechanisms, such as the regularizing effect resulting from minimizing appropriately averaged log-likelihood functions~\cite{Maaten2013,wei2020implicit}, bagging or feature corruption where the input data is being perturbed~\cite{Scholkopf1997,Maaten2013,Chen2012}, or as maximizing a lower bound on the Bayesian marginal likelihood~\cite{Wang2013}.

However, the most natural interpretation of dropout is that dropout provides a form of Bayesian model averaging, in which the dropout-perturbations induces a distribution over the deterministic neural networks predictions. In this view, the role of an interpretation of dropout is to describe this distribution. One such interpretation was given in the seminal work \cite{gal2016dropout}, in which the posterior of a variant of dropout (Monte-Carlo dropout) is shown to be an approximation of a deep Gaussian process. 

This work takes a more direct approach, namely to identify the circumstances under which dropout provides the unique optimal assignment of degrees of rational belief. That is, instead of describing the effect of dropout on a model (for instance, as inducing a Bayesian model averaging procedure through the perturbations), we view dropout as a specific model invariance, and the resulting distribution --i.e., Bayesian model with dropout enabled-- as simply the maximum entropy distribution under this invariance. For this reason, we dub the method \emph{Bayesian dropout}.

We accomplish this by using that Bayesian inference is a special case of the principle of maximum entropy (ME)~\cite{Caticha2006}, which naturally allows us to formulate dropout as a constraint on the space of models within which we are maximizing the entropy. As this view is model-agnostic, dropout can be applied as a principled tool in Bayesian modeling to any probabilistic model for which a dropout-perturbation can be defined. We illustrate the method using a Bayesian linear regression and when maximizing the resulting likelihood function we recover non-Bayesian dropout approaches for linear regression based on loss-functions~\cite{Wang2013,Maaten2013}.

\section{Methods}
The goal of machine learning is to predict, explain and control the environment in a rational manner based on new information and past beliefs. R. T. Cox showed that a theory of degrees of rational beliefs is only consistent if they obey the rules of probability theory~\cite{Cox1946}, and this insight has given rise to Bayesian methods, where the past beliefs are identified with priors on the model parameters, and the available information is the observed data and the relationship between data $\m x$ and parameters $\theta$. This leads to Bayes' theorem
$p(\m \theta | \m x ) = \frac{p(\m x | \m \theta) p(\m \theta)}{p(\m x)}$.  
However, Bayes' theorem is not the most general way of arriving at rational beliefs. In many situations the relevant information available in the form of constraints, for instance an ideal gas where the relevant constraints include energy conservation. In this case, 
the method of \emph{maximum entropy}, MaxEnt, allows assignment of rational beliefs under expectation constraints~\cite{Jaynes1957}. 

A more general system of rational inference must be able to handle information in the form of both observed data and arbitrary constraints in an objective manner. 
If these constraints are available to us in the form of a likelihood (a constraint on the class of posterior functions) and observed data, the method must reduce to Bayesian inference. If the constraint is in the form of expectations, it must reduce to MaxEnt. This can be accomplished by the \emph{extended method of maximum entropy} (ME), which contain both methods as special cases~\cite{Caticha2006}.

\paragraph{Extended method of maximum entropy:}
Let $\m z$ be all the variables under consideration. The goal for a rational learner is to update from a prior distribution $q(\m z)$ to a posterior distribution $p(\m z)$ when new information is made available.

The relevant information can come in the form of observed data, priors, the form of the likelihood or expectations, all of which constrain the posterior to belong to a family of distributions $p \in \mathcal{C}$. The method of ME makes the assumption that not all distributions $p \in \mathcal{C}$ are equally desirable, and assume they can be put in an order of preference represented by the functional $S[p,q]$. Using the further assumptions of locality, coordinate invariance, consistency for independent subsystems, it can be shown that $S$ must have the form~\cite{Caticha2006}
\begin{equation}
	S[p,q] = \intd{d\m z} p(\m z)\log\frac{p(\m z)}{q(\m z)}. \label{eqn:S}
\end{equation}
In a machine-learning context, we would consider $\m z$ as consisting of  $n$ observed data points $\m x_i$ and parameters of interest $\m \theta$. In this case, the relevant information is that we observed an actual value of the data, $\m x'$, and this places the constraint on the family of posteriors $p$ that they must belong to
\begin{equation}
	\mcal C  = \left\{ p \mid \smallint\! d\m \theta \ p(\m x, \m \theta) = \delta(\m x - \m x') \right\}, \label{eqn:C}
\end{equation}
where $\delta$ is the delta-function. Maximizing \cref{eqn:S} under this constraint exactly recovers Bayes' theorem~\cite{Caticha2006}. 

\paragraph{Bayesian Dropout}
The neural network formulation of dropout (c.f. \cite{Hinton2012}) consider a function $\m y = f_{\m \theta}(\m x)$ (the neural network) parameterized by a set of weights $\m \theta$ which maps inputs to outputs. The simplest way to train the network is by gradient descent with respect to $\m \theta$ on the loss function averaged over input and output training pairs. Dropout is a simple extension where between each gradient descent step and for each observation $\m x_i$ a perturbed set of parameters is generated $\m \thetat_i \sim p(\cdot | \m \theta)$
and a single gradient update is performed on the modified error function
$E(\m \theta) = \sum_i L(y_i,  f_{\m \thetat_i}(x_i)) )$.

The perturbed set of parameters are most commonly obtained by simply blanking out a fraction of the hidden units independently~\cite{Hinton2012}. In the limit of low learning rate on $\m \theta$, dropout will favor weights $\m \theta$ that tend to give good performance when a fraction of the inputs are missing thereby reducing co-adaptation~\cite{Hinton2012}.
The main difference between dropout and simple model averaging is that dropout does not weight a given set of perturbed parameters by the posterior probability, and it is for this reason we need to specify dropout using the ME framework.

Consider a Bayesian equivalent of a neural network model with joint likelihood $p(\m x, \m \theta) = p(\m x | \m \theta)p(\m \theta)$ where the likelihood term, for $i=1,\dots,n$, is
$y_i | \m \theta  \sim \mathrm{Normal}(\cdot \ ; \ f_{\m \theta}(x_i), \sigma^2)$.
It is easy to see that assuming flat priors the MAP solution to this model will be exactly equivalent to the global maximum of the neural-network error. We might consider adding a dropout step to the generative model so it becomes:
\begin{equation}
	\m \theta  \sim p(\cdot), \quad \quad
	\mbox{for each $i$: } \quad \m \thetat_i   \sim p(\cdot | \m \theta), \quad \quad
	y_i | \m \thetat  \sim \mathrm{Normal}(\cdot \ ; \ f_{\m \thetat_i }(x_i), \sigma^2). \label{dropout_generative}
\end{equation}
Although this generative process bears some similarity to dropout, it will infer which weights are best to blank out to explain each observation, and \emph{not} describe a parameter specification which is robust to \emph{having} parameters blanked out. The effect of this will be twofold. (i) It will create a mixture of models, each model (corresponding to a set of dropped-out parameters) being allowed to co-adapt to the data. (ii) the mixture will be weighted with the posterior probability, thereby reducing the effective number of components.

To implement dropout, we must specify dropout as an external constraint on the learners representation, namely that individual parameters are zeroed stochastically. To put this in words, \emph{Bayesian Dropout} can be defined as \emph{the constraint that the dropout distribution is not inferred from data}.
To implement this using ME, we first need to fix the prior measure $q$ and the relevant constraints. 
As prior distribution $q$ we adopt the same functional form as the naive Bayesian dropout \cref{dropout_generative}. Using $\m \thetat = (\m \thetat_i)_{i=1}^n$, the joint distribution may be written 
\begin{equation}
	q(\m x, \m \thetat, \m \theta) = q(\m x | \m \thetat) q(\m \thetat | \m \theta) q(\m \theta).
\end{equation}

Next we specify the constraints: The first constraint is simply the data-constraint that we observed an actual value of $\m x$, namely $\m x'$ (see \cref{eqn:C}). The second constraint is the dropout-condition. To say weights are dropped out stochastically, and that this distribution is not inferred from data, is saying exactly that the dropout weights must depend on $\m \theta$, but \textbf{not} on $\m x$. We can therefore identify Bayesian Dropout with the constraint
\begin{equation}
	p(\m \thetat | \m x, \m \theta) \equiv q(\m \thetat | \m \theta) \ \quad \textit{(Bayesian Dropout)} \label{eqn:bdcondition}
\end{equation}
Accordingly we have $p(\m x, \m \thetat, \m \theta) = p(\m \thetat | \m \theta) p(\m x, \m \theta)$. The ordering  \cref{eqn:S} becomes:
\begin{equation}
	S[p,q]  = \intd{d\m x\m d\thetat d\m \theta} p(\m x, \m \thetat, \m \theta) \log \frac{  p(\m x, \m \thetat, \m \theta)} { q(\m x, \m \thetat, \m \theta)} \nonumber 
	= \intd{d\m x\m d\thetat d\m \theta} p(\m x, \m \theta)p(\m \thetat | \m \theta) \log \frac{  p(\m x, \m \theta)} { q(\m x | \m \thetat)q(\m \theta)}. \label{eqn:varproblem}
\end{equation}
The distribution which uniquely maximizes this functional can be found by taking the functional derivative with respect to $p(\m x, \m \theta)$ while introducing lagrange multipliers $\lambda_{\m x}$ to handle the (infinite) number of data-constraints \cref{eqn:C} and $\alpha$ to handle the sum constraint. This leads to the variational problem:
\begin{equation}
	0 = \frac{\delta }{\delta p(\m x,\m \theta)}  \left\{ S[p,q] + \alpha\left[\intd{d\m x d\m \thetat d \m \theta} p(\m x,\m \thetat,\m \theta) - 1\right] + \int d\m x \lambda_{\m x}\left[\intd{d\m \thetat d\m \theta} p(\m x,\m \thetat, \m \theta) - \delta(\m x-\m x') \right]  \right\}.
\end{equation}
Performing the functional derivative and solving for $p(\m x, \m \theta)$ gives
\begin{equation}
	p(\m x, \m \theta) = \frac{1}{z} q(\m \theta) \exp\left( \intd{d \m \thetat} q(\m \thetat | \m \theta) \log q(\m x | \m \thetat) + \lambda_{\m x}\right), \label{eqn:plambda}
\end{equation}
where $z$ handles normalization. Recall the requirement that the posterior is consistent with the observed data $\m x'$
$\smallint \! d\m \theta\  p(\m x, \m \theta) = \delta(\m x - \m x')$.
Using this identity on the right-hand side of \cref{eqn:plambda} to fix the Lagrange multipliers $\lambda_{\m x}$
\begin{equation}
	p(\m x, \m \theta)  = \frac{1}{\mcal Z(\m x)} q(\m \theta) \exp\left( \intd{d \m \thetat} q(\m \thetat | \m \theta) \log q(\m x | \m \thetat) \right)\delta(\m x - \m x'), \quad
	\mcal Z(\m x)   = \intd{d \m \theta}q(\m \theta) \exp \intd{d \m \thetat} p(\m \thetat | \m \theta) \log q(\m x | \m \thetat),
\end{equation}
such that the posterior distribution of the parameters becomes
\begin{equation}
	p(\m \theta)  = \frac{1}{Z(\m x')} q(\m \theta) \exp\left( \intd{d \m \thetat} q(\m \thetat | \m \theta) \log q(\m x' | \m \thetat) \right). \label{eqn:bd}
\end{equation}
Note that $p$ refers to our final state of knowledge and is therefore not conditioned on $\m x'$.  If $p(\m \thetat | \m \theta) = \delta(\m \thetat - \m \theta)$ the expression reduces to Bayes' theorem, but otherwise values of $\m \theta$ which are not robust to dropout will be penalized by the log term. 

\paragraph{Bayesian linear regression with dropout:} 
We examine Bayesian dropout by applying it to a conjugated Bayesian linear regression model~\cite{Gelman2003}. The joint prior measure $q$ is defined by the generative process
\begin{equation}
	\sigma^2  \sim \textrm{Gamma}(a_0,b_0), \\
	\m w | \sigma^2  \sim \Normal\big(\m 0, \frac{\sigma^2}{\lambda_0}\m I  \big), \\
	\m \wt_i |\m w  \sim p_f(\m w), \\
	y_i | \m \wt_i,\sigma^2  \sim  \Normal(\m x_i^T \m \wt_i, \sigma^{2}),
\end{equation}\label{eqn:blr}
where $\m y$ is an $n$-dimensional vector of responses, and $\m w$ and
$\m x_i$ are $p$-dimensional vectors of weights and covariates. Compared to \cref{eqn:bd}, $\m \theta = (\m w,\sigma)$.  We consider the limit $a_0,b_0 \rightarrow 0$ corresponding to the Jeffreys prior $\sigma^2 \sim \sigma^{-2}$. As a dropout distribution $p_f(\m w)$, we consider independent binary dropout with rate $f$,
\begin{equation}
	\label{eq:dropout_distribution}
	p_f(\m w) = \prod_{d=1}^p [ (1-f) \delta(\wt_{id} - w_{d}) + f \delta(\wt_{id})].
\end{equation}
Computing the expectation of the log likelihood with respect to the dropout distribution yields
\begin{equation}
	\big\langle \log q(\m y | \m \wt)\big\rangle = -\frac{n}{2}\log(2
	\pi \sigma^2) - \frac{1}{2\sigma^2} (\m y - \hat{\m y})^\top (\m y - \hat{\m y})
	- f(1-f)\frac{1}{2\sigma^2} \m w^T \m \Lambda_1 \m w,
\end{equation}
where  $\hat{\m y} = (1-f)\m X \m w$, $\m \Lambda_1 = (\m X^\top \m X) \circ \m I$, $\m X = [\m x_1, \m x_2, \cdots, \m x_n]^T$ is an $n \times p$ matrix of covariates and 
$\circ$ is the Hadamard product. This is combined with the conjugate prior for $\m w$,
$p(\m w|\sigma^2) = \left(\frac{\lambda_0}{2\pi\sigma^2}\right)^{p/2} \exp\left(-\frac{\lambda_0}{2\sigma^2}\m w^\top \m w\right)$, gives the posterior
\begin{equation}
	p(\m w, \sigma | \m y) =
	\frac{1}{(2\pi\sigma^2)^\frac{n+p}{2}}\exp\left(-\frac{1}{2\sigma^2}\left[(\m w - \m \mu_n)^T\m \Lambda(\m w - \m \mu_n) + \m y^T \m y - \m \mu_n^T \m \Lambda \m \mu_n \right] \right), \label{eq:BLRPosterior}
\end{equation}
where 
$\m \Lambda = \lambda_0\m I + f(1-f)\m \Lambda_1 +(1-f)^2\m X^T \m X$ and $ \m \mu_n = (1-f)\m \Lambda^{-1}\m X^T \m y$.
Examining \cref{eq:BLRPosterior}, we see this is equal to the familiar Bayesian linear model in the limit $f=0$, 
but more generally Bayesian dropout serves as a data dependent prior. The weights of larger features are regularized more heavily by dropout than by ridge regression as also observed by \cite{Wang2013}. Also note that removing the prior term $\lambda_0$ and taking the maximum likelihood of eq.~\ref{eq:BLRPosterior} with respect to $\m w$ while keeping $\sigma$ constant recover the expression for learning with marginalized corrupted features~\cite{Maaten2013}.
\begin{figure}
	\centering
	\includegraphics[width=0.36\linewidth]{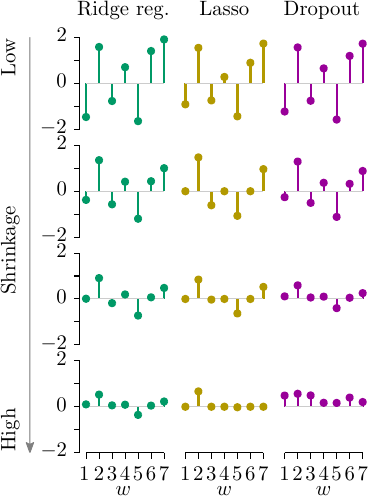}\hspace{1.5cm}
	\includegraphics[width=0.36\linewidth]{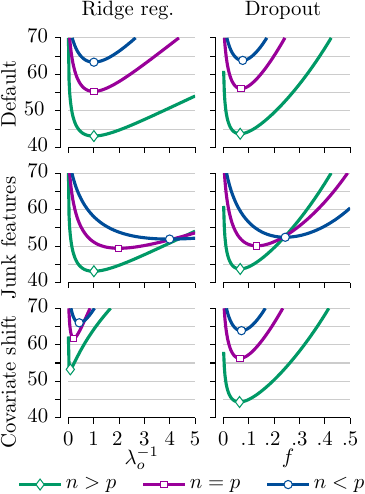}
	\caption{(a) Comparison of dropout, ridge regression and Lasso for the quantitative structure property relationship (QSPR) data. (b) Performance of ridge regression and dropout in the three scenarios and three conditions.}
	\label{fig:toydata}
\end{figure}

\paragraph{Dropout as parameter shrinkage}
Bayesian priors often understood as improving performance through regularization. From eq.~(\ref{eq:BLRPosterior}) it is evident that in the linear regression model, when the prior is $\lambda_0=0$, dropout shrinks parameters towards $\m \Lambda_1^{-1}\m X^\top \m y$, corresponding to the maximum likelihood estimate of each weight in isolation of the other (due to the Hadamard product in $\m \Lambda_1$). Thus, dropout provides shrinkage towards a solution with no co-adaptation. This is illustrated in \cref{fig:toydata} (a) for a quantitative structure property relationship (QSPR) data example~\cite{Wold2001} with $n=19$ observation and $p=7$ highly correlated covariates. The sign of the DGR feature was flipped so that all covariates were positively correlated with the response. In contrast ridge regression and Lasso, dropout shrinks the parameters towards an all positive solution, as would be expected when all covariates are positively correlated with the response.

In \cref{fig:toydata} (b), 
we examined the generalization performance of Baysian dropout under model misspecification. We generated $n=20$ training and test data from the normal linear regression model with $\lambda_0=\sigma^2=1$ and plotted the crossvalidation squared error averaged over 100\,000 random data sets. In dropout we used an incorrect prior $\lambda_0=10^{-3}$ to examine its performance under model misspecification. The covariates $\m X$ were chosen as $\m X = \m R \m L$ where $\m R$ was a $20\times10$ standard normal i.i.d. random matrix and $\m L$ was a $10\times p$ random projection matrix where each column had unit length. We considered both the underdetermined ($n>p=10$), determined ($n=p=20$), and overdetermined ($n<p=40$) scenario under three different conditions: In the \emph{default} condition, the response was generated directly from the model. In this condition, ridge regression with the correct prior is optimal, and dropout was found to perform on par. In the \emph{junk features} condition, we set all but the first $10$ weights to zero when generating the data, corresponding to a misspecified prior or as having $p-10$ noninformative covariates. Again performing on par, both ridge regression and dropout could counter this by increasing the amount of regularization. Finally, in the \emph{covariate shift} condition we generated the data as in the default condition but multiplied each covariate by a normal random number before fitting the models. We found that ridge regression was quite sensitive to this type of model mismatch whereas dropout was significantly more robust.

\begin{figure}
	\centering
	\includegraphics[width=0.85\linewidth]{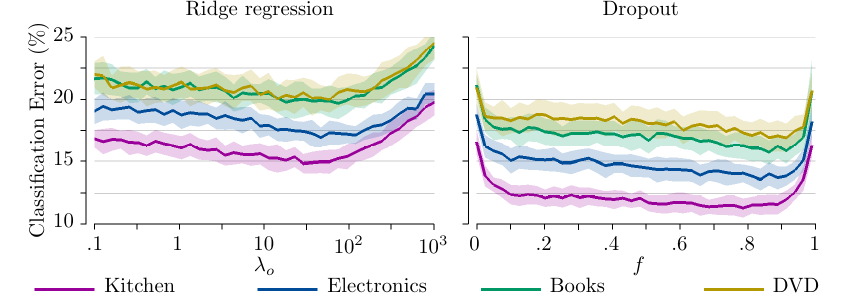}
	\caption{Performance of Bayesian linear regression with $L_2$
		regularization (ridge regression) and dropout on four binary
		prediction problems.}
	\label{fig:amazon}
\end{figure}

Finally, we evaluated Bayesian linear regression with dropout on the
Amazon review dataset~\cite{blitzer-dredze-pereira:2007:ACLMain}. The
dataset consists of a bag-of-words representation of product reviews
in four categories: Books, DVDs, Electronics, and Kitchen, and the
task is to predict if the review is positive or negative. The number
of observations ranges from 3\,587 to 5\,946 and the number of
covariates from 123\,099 to 193\,220. We randomly selected 75\% of the
documents as training data and computed the predictive error on the
remaining documents (see Figure~\ref{fig:amazon}). We computed the
average and standard deviation of the test error over 40 simulations
for each data point for Bayesian linear regression with and without
dropout for varying values of the scale parameter $\lambda_0$. Dropout was found to significantly improved generalization performance for a wide
range of parameter settings.

\section{Discussion and conclusion}
A classical result first proved by J. L. Doob in 1948 is that for finite parameter spaces the Bayesian posterior distribution will almost surely concentrate on the true parameter value provided a consistent estimator exists and the true parameter value is in support of the prior~\cite{MR0033460}. In this light it might be surprising that additionally restricting the posterior model class can provide any benefits; however, these results must be interpreted in the light of model misspecification. 

When the model is misspecified, Bayesian learning will concentrate the posterior around values of $\m \theta$ with the highest likelihood, however, depending on how the model is misspecified, there are no additional guarantees, and the posterior can converge to parameter values which do not give the best prediction~\cite{Grunwald2007,gelman2012philosophy}. 
This situation is similar to what motivated dropout in neural networks, and may provide reasons to add constraints such as dropout to make the model less likely to co-adapt.

This also sheds some light on the more general question where constraints come from in the first place. The model and prior probabilities commonly reflect a tradeoff between convenience, scientific knowledge, and tractability~\cite{gelman2012philosophy}, and under this view, Bayesian dropout can be well motivated as either reflecting scientific knowledge such as energy conservation, or the simple fact the model is \emph{known} to be misspecified in such a way it is desirable to prevent co-adaptation by adding an appropriate constraint.

Dropout provides a simple yet powerful tool to avoid co-adaptation in neural networks and has been shown to offer tangible benefits; however, its formulation as an algorithm rather than as a set of probabilistic assumptions precludes its use in Bayesian modelling. We have shown how dropout can be interpreted as optimal inference under a particular constraint. This qualifies dropout beyond being a particular optimization procedure, and has the advantage of giving researchers who want to apply dropout to a particular model a principled way to do so.

We have demonstrated Bayesian dropout on an analytically tractable regression model, providing a probabilistic interpretation of its mechanisms for regularizing and preventing co-adaptation as well as its connection to other Bayesian techniques. In our experiments we find that dropout can provide robustness under model misspecification, and offer benefits over ordinary Bayesian linear regression in a real dataset.

\bibliographystyle{elsarticle-harv}

\bibliography{library}

\end{document}